\documentclass[11pt,a4paper]{article}
\usepackage[hyperref]{acl2018}
\usepackage{times}
\usepackage{latexsym}

\usepackage{url}

\aclfinalcopy 
\def\aclpaperid{868} 

\usepackage{xcolor,colortbl}
\usepackage{multirow}
\usepackage[utf8]{inputenc} 
\usepackage[T1]{fontenc}    
\usepackage{url}            
\usepackage{booktabs}       
\usepackage{amsfonts}       
\usepackage{nicefrac}       
\usepackage{microtype}      
\usepackage{amsmath}
\usepackage{graphicx}
\usepackage{subcaption}
\usepackage{todonotes}

\makeatletter
\def\BState{\State\hskip-\ALG@thistlm}
\makeatother

\aclfinalcopy 
\def\aclpaperid{868} 

\title{Bleaching Text: Abstract Features for Cross-lingual Gender Prediction }

\author
{
	\begin{tabular}{ccccc}
	Rob van der Goot$^{\heartsuit}$ & Nikola Ljubešić$^{\spadesuit}$ & Ian Matroos$^{\heartsuit}$ & Malvina Nissim$^{\heartsuit}$ & Barbara Plank$^\heartsuit$$^{\clubsuit}$ \\
	\end{tabular}
	\\
    $^\heartsuit$ Center for Language and Cognition, University of Groningen, The Netherlands\\
    $^\spadesuit$ Department of Knowledge Technologies,
Jozef Stefan Institute, Ljubljana, Slovenia\\
    $^\clubsuit$ IT University of Copenhagen, Copenhagen, Denmark\\
	{\tt \small{\{r.van.der.goot,i.matroos,m.nissim\}@rug.nl,nljubesi}@gmail.com,bplank@itu.dk}
}

\date{}

\begin{document}
\maketitle
\begin{abstract}
Gender prediction has typically focused on lexical and social network features, yielding good performance, but making systems highly language-, topic-, and platform-dependent. Cross-lingual embeddings	 circumvent some of these limitations, but capture  gender-specific style less. 
We propose an alternative: \textit{bleaching text}, i.e., transforming lexical strings into more abstract features. This study provides evidence that such features allow for better transfer across languages. Moreover, we present a first study on the ability of humans to perform \textit{cross-lingual} gender prediction. We find that human predictive power proves similar to that of our bleached models, and both perform better than lexical models.
\end{abstract}

\section{Introduction}

\textit{Author profiling} is the task of discovering latent user attributes disclosed through text, such as gender, age, personality, income, location and occupation~\cite{rao2010classifying,burger11-discriminating,feng2012,jurgens2013,bamman14-gender,plank-hovy:2015:WASSA,flekova:etal:2016}. It is of interest to several applications including personalized machine translation, forensics, and marketing~\cite{mirkin2015,rangel2015overview}.

Early approaches to gender prediction \cite[e.g.]{koppel2002,schler06-effects} are inspired by pioneering work on authorship attribution~\cite{mosteller1964inference}. Such stylometric models typically rely on carefully hand-selected sets of content-independent features to capture style beyond  topic. Recently, \textit{open vocabulary} approaches~\cite{schwartz2013personality}, where the entire linguistic production of an author is used, yielded substantial performance gains in online user-attribute prediction~\cite{nguyen2014gender,preoctiuc2015occupation,emmery-chrupala-daelemans:2017:WNUT}. Indeed, the best performing gender prediction models exploit chiefly lexical information~\cite{rangel2017overview,basile2017pan}.

Relying heavily on the lexicon though has its limitations, as it results in models with limited portability. Moreover, performance might be overly optimistic due to topic bias~\cite{sarawgi2011gender}.  
Recent work on cross-lingual author profiling has proposed the use of solely language-independent features~\cite{ljubevsic:ea:2017}, e.g.,  specific textual elements (percentage of emojis, URLs, etc) and users' meta-data/network (number of followers, etc), but this information is not always available. 

We propose a novel approach where the actual text is still used, but bleached out and transformed into more abstract, and potentially better transferable features. One could view this as a method in between the open vocabulary strategy and the stylometric approach. It has the advantage of fading out content in favor of more shallow patterns still based on the original text, without introducing additional processing such as part-of-speech tagging. In particular, we investigate to what extent gender prediction can rely on generic non-lexical features (\textbf{RQ1}), and  how predictive such models are when transferred to other languages (\textbf{RQ2}). We also glean insights from human judgments, and investigate how well people can perform cross-lingual gender prediction (\textbf{RQ3}). 
We focus on gender prediction for Twitter, 
motivated by data availability.

\paragraph{Contributions} In this work i) we are the first to study cross-lingual gender prediction without relying on users' meta-data; ii) we propose a novel simple abstract feature representation which is surprisingly effective; and iii) we gauge human ability to perform cross-lingual gender detection, an angle of analysis which has not been studied thus far.

\section{Profiling with Abstract Features}
Can we recover the gender of an author from bleached text, i.e., transformed text were the raw lexical strings are converted into abstract features? We investigate this question by building a series of predictive models to infer the gender of a Twitter user, in absence of additional user-specific meta-data.  
Our approach can be seen as taking advantage of elements from a data-driven open-vocabulary approach, while trying to capture gender-specific style in text beyond topic.

To represent utterances in a more language agnostic way, we propose to simply transform the text into alternative textual representations, which deviate from the lexical form to allow for abstraction. 
We propose the following transformations, exemplified in Table~\ref{tab:exampleSents}. They are mostly motivated by intuition and inspired by prior work, like the use of shape features from NER and parsing~\cite{petrov-klein:2007:main,schnabel2014flors,plank2016ea,limsopatham-collier:2016:WNUT}:

 \begin{table}\centering
 \resizebox{\columnwidth}{!}{%
 	\begin{tabular}{l l l l l l l l l}
     \toprule
         \textbf{Original}  	& a  & bag 	& of	& Doritos& for	& lunch!& \includegraphics[width=.3cm]{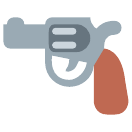}\includegraphics[width=.3cm]{gun}\includegraphics[width=.3cm]{gun}\includegraphics[width=.3cm]{gun} \\
         \textbf{Frequency} 	& 4  & 2   	& 4  	& 0 	 & 4 	& 1 	& 0\\
		\textbf{Length} 		& 01 & 03  	& 02  	& 07 	 & 03 	& 06 	& 04\\ 
         \textbf{PunctC}		& W  & W 	& W 	& W 	 & W	& W!    & \includegraphics[width=.3cm]{gun}\includegraphics[width=.3cm]{gun}\includegraphics[width=.3cm]{gun}\includegraphics[width=.3cm]{gun}\\
         \textbf{PunctA}		& W  & W 	& W 	& W 	 & W 	& WP    & JJJJ \\
         \textbf{Shape}			& L  & LL  	& LL 	& ULL	 & LL	& LLX   & XX\\
         
         \textbf{Vowels}	    & V  & CVC 	& VC 	& CVCVCVC& CVC	& CVCCCO& OOOO \\

         \bottomrule
 	\end{tabular}
     } %
     \caption{Abstract features example transformation.}
     \label{tab:exampleSents}
 \end{table}

\begin{itemize}

\item \textbf{Frequency} Each word is presented as its binned frequency in the training data; bins are sized by orders of magnitude.
\item \textbf{Length} Number of characters (prefixed by 0 to avoid collision with the next transformation). 

\item \textbf{PunctC} Merges all consecutive alphanumeric characters to one `W' and leaves all other characters as they are (C for conservative).  

\item \textbf{PunctA} Generalization of PunctC (A for aggressive), converting different types of punctuation to classes:  emoticons\footnote{Using the NLTK tokenizer \url{http://www.nltk.org/_modules/nltk/tokenize/casual.html}} to `E' and emojis\footnote{\url{https://pypi.python.org/pypi/emoji/}} to `J', other punctuation to `P'. 

\item \textbf{Shape} Transforms uppercase characters to `U', lowercase characters to `L', digits to `D' and all other characters to `X'. Repetitions of transformed characters are condensed to a maximum of 2 for greater generalization.

\item  \textbf{Vowel-Consonant} To approximate vowels, while being able to generalize over (Indo-European) languages, we convert any of the `aeiou' characters to `V', other alphabetic character to `C', and all other characters to `O'. 
\item \textbf{AllAbs}  A combination (concatenation) of all previously described features. 

\end{itemize}

\section{Experiments}

In order to test whether abstract features are effective and transfer across languages, we set up experiments for gender prediction comparing lexicalized and bleached models for both in- and cross-language experiments. We compare them to a model using multilingual embeddings~\cite{ruder2017survey}. Finally, we elicit human judgments both within language and across language. The latter is to check 
whether a person with no prior knowledge of (the lexicon of) a given language can predict the gender of a user, and how that compares to an in-language setup and the machine. If humans can predict gender cross-lingually, they are likely to rely on aspects beyond lexical information.

\paragraph{Data}
We obtain data from the \textsc{TwiSty} corpus~\cite{twisty}, a multi-lingual collection of Twitter users, for the languages with 500+ users, namely Dutch, French, Portuguese, and Spanish. 
We complement them with English, using data from a predecessor of \textsc{TwiSty}~\cite{plank-hovy:2015:WASSA}. All datasets contain manually annotated gender information. 
To simplify interpretation for the cross-language experiments, we balance gender in all datasets by downsampling to the minority class. 
The datasets' final sizes are given in Table~\ref{tbl:resultsMain}. We use 200 tweets per user, as done by previous work~\citep{twisty}. We leave the data untokenized to exclude any language-dependent processing, because original tokenization could preserve some signal. Apart from mapping usernames to `USER' and urls to `URL' we do not perform any further data pre-processing.

\begin{table*}[htb!]\centering
\resizebox{\textwidth}{!}{%

\begin{tabular}{l|r|c|c|cc|c|cc}
\toprule
&  & \multicolumn{2}{c|}{\sc in-language} & \multicolumn{5}{c}{\sc cross-language} \\ \cline{3-9}
\raisebox{.5em}{\textsc{Test}}  & \raisebox{.5em}{\textsc{Users}} & \textsc{Lexical} & \textsc{Abstract} & \textsc{Lex Avg} & \textsc{Lex All} & \textsc{Embeds} & \textsc{Abs Avg} & \textsc{Abs All}  \\
  \midrule
EN &   850 & 69.3 & 66.1 & 51.8 & 50.5 & 61.6 & 55.3 & 59.8 \\
NL &   894 & 81.3 & 71.8 & 52.3 & 50.0 & 56.8 & 59.5 & 69.2 \\
FR & 1,008 & 80.8 & 68.3 & 53.4 & 53.8 & 50.0 & 58.7 & 65.4 \\
PT & 3,066 & 86.0 & 68.1 & 55.3 & 63.8 & 59.5 & 59.3 & 58.9 \\
ES & 8,112 & 85.3 & 69.8 & 55.6 & 63.5 & 71.3 & 56.6 & 66.0 \\

\bottomrule
\end{tabular}
}
\caption{Number of users per language and results for gender prediction (accuracy). \textsc{In-Language}: 10-fold cross-validation. \textsc{Cross-Language}: Testing on all test data in two setups: averages over single source models (\textsc{Avg}) or training a single model on all languages except the target  (\textsc{All}). Comparison of lexical n-gram models (\textsc{Lex}), bleached models (\textsc{Abs}) and multilingual embeddings model (\textsc{Embeds}).}
\label{tbl:resultsMain}
\end{table*}

\subsection{Lexical vs Bleached Models}
\label{sec:models}
We use the \texttt{scikit-learn}~\cite{scikit-learn} implementation of a linear SVM with default parameters (e.g., L2 regularization). We use 10-fold cross validation for all in-language experiments. For the cross-lingual experiments, we train on all available source language data and test on all target language data. 

For the lexicalized experiments, we adopt the features from the best performing system at the latest PAN evaluation campaign\footnote{\url{http://pan.webis.de}} \cite{basile2017pan} (word 1-2 grams and character 3-6 grams). 

For the multilingual embeddings model we use the mean embedding representation from the system of ~\cite{plank2017all} and add max, std and coverage features. We create multilingual embeddings by projecting monolingual embeddings to a single multilingual space for all five languages using a recently proposed SVD-based projection method with a pseudo-dictionary~\cite{smith2017}. 
The monolingual embeddings are trained on large amounts of in-house Twitter data (as much data as we had access to, i.e., ranging from 30M tweets for French to 1,500M tweets in Dutch, with a word type coverage between 63 and 77\%). This results in an embedding space with a vocabulary size of 16M word types. All code is available at~\url{https://github.com/bplank/bleaching-text}.  

For the bleached experiments, we ran models with each feature set separately. In this paper, we report results for the model where all features are combined, as it proved to be the most robust across languages. 
We tuned the $n$-gram size of this model through in-language cross-validation, finding that $n=5$ performs best.  

 When testing across languages, we report  accuracy for two setups: average accuracy over each single-language model (\textsc{Avg}), and accuracy obtained when training on the concatenation of all languages but the target one (\textsc{All}). The latter setting is also used for the embeddings model. We report accuracy for all experiments.

\begin{table}[h!]\centering
\resizebox{\columnwidth}{!}{%
\begin{tabular}{l cccccc} 
   \toprule
& \textbf{Test $\rightarrow$}  & EN & NL & FR & PT & ES \\
   \midrule
\multirow{5}{*}{\rotatebox[origin=c]{90}{\textbf{Train}}} 
& EN &      & 52.8 & 48.0 & 51.6 & 50.4 \\
& NL & 51.1 &      & 50.3 & 50.0 & 50.2 \\
& FR & 55.2 & 50.0 &      & 58.3 & 57.1 \\
& PT & 50.2 & 56.4 & 59.6 &      & 64.8 \\
& ES & 50.8 & 50.1 & 55.6 & 61.2 &      \\
\midrule
& Avg & 51.8  & 52.3  & 53.4  & 55.3  & 55.6 \\
\bottomrule
\end{tabular}
}
\caption{Pair-wise results for lexicalized models.}
\label{tbl:pairwise}
\end{table}

\paragraph{Results and Analysis}

Table~\ref{tbl:resultsMain} shows results for both the cross-language and in-language experiments in the lexical and abstract-feature setting.

Within language, the lexical features unsurprisingly work the best, achieving an average accuracy of 80.5\% over all languages. The abstract features lose some information and score on average 11.8\% lower, still beating the majority baseline (50\%) by a large margin (68.7\%). If we go across language, the lexical approaches break down (overall to 53.7\% for \textsc{Lex Avg}/56.3\% for \textsc{All}), except for Portuguese and Spanish, thanks to their similarities (see Table~\ref{tbl:pairwise} for pair-wise results). The closely-related-language effect is also observed when training on all languages, as scores go up when the classifier has access to the related language. The same holds for the multilingual embeddings model. On average it reaches an accuracy of 59.8\%.

The closeness effect for Portuguese and Spanish can also be observed in language-to-language experiments, where scores for ES$\mapsto$PT and PT$\mapsto$ES are the highest.
Results for the lexical models are generally lower on English, which might be due to smaller amounts of data (see first column in Table~\ref{tbl:resultsMain} providing number of users per language).

The abstract features fare surprisingly well and work a lot better across languages. The performance is on average 6\% higher across all languages (57.9\% for \textsc{Avg}, 63.9\% for \textsc{All}) in comparison to their lexicalized counterparts, where \textsc{Abs All} results in the overall best model. For Spanish, the multilingual embedding model clearly outperforms \textsc{Abs}. However, the approach requires large Twitter-specific embeddings.\footnote{We tested the approach with more generic (from Wikipedia) but smaller (in terms of vocabulary size) Polyglot embeddings resulting in inferior multilingual embeddings for our task.}

\begin{table}
\centering
\begin{tabular}{r r r}
\toprule
& Male & Female \\
\midrule
1 & W W W W "W" & USER E W W W \\
2 & W W W W ? & 3 5 1 5 2 \\
3 & 2 5 0 5 2 & W W W W \includegraphics[width=.3cm]{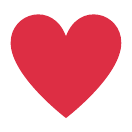}  \\ 
4 & 5 4 4 5 4 & E W W W W \\
5 & W W, W W W? & LL LL LL LL LX \\
6 & 4 4 2 1 4 & LL LL LL LL LUU \\
7 & PP W W W W & W W W W *-* \\
8 & 5 5 2 2 5 & W W W W JJJ \\
9 & 02 02 05 02 06 & W W W W \&W;W \\
10 & 5 0 5 5 2 & J W W W W \\
\bottomrule
\end{tabular}
\caption{Ten most predictive features of the \textsc{Abs} model across all five languages. Features are ranked by how often they were in the top-ranked features for each language. Those prefixed with 0 (line 9) are length features. The prefix is used to avoid clashes with the frequency features.}
\label{tbl:features}
\end{table}

For our \textsc{Abs} model, if we investigate predictive features over all languages, cf.\ Table~\ref{tbl:features}, we can see that the use of an emoji (like \includegraphics[scale=0.2]{heart}) and shape-based features are predictive of female users. Quotes, question marks and length features, for example, appear to be more predictive of male users.

\subsection{Human Evaluation}

We experimented with three different conditions, one within language and two across language. For the latter, we set up an experiment where native speakers of Dutch were presented with tweets written in Portuguese and were asked to guess the poster's gender. In the other experiment, we asked speakers of French to identify the gender of the writer when reading Dutch tweets. In both cases, the participants declared to have no prior knowledge of the target language. For the in-language experiment, we asked Dutch speakers to identify the gender of a user writing Dutch tweets. The Dutch speakers who participated in the two experiments are distinct individuals. Participants were informed of the experiment's goal. Their identity is anonymized in the data.

We selected a random sample of 200 users from the Dutch and Portuguese data,  preserving a 50/50 gender distribution. Each user was represented by twenty tweets. The answer key (F/M) order was randomized. 
For each of the three experiments we had six judges, balanced for gender, and obtained three annotations per target user. 



\paragraph{Results and Analysis}

Inter-annotator agreement for the tasks was measured via Fleiss kappa ($n=3, N=200$), and was higher for the in-language experiment ($K=0.40$) than for the cross-language tasks (NL$\mapsto$PT: $K=0.25$; FR$\mapsto$NL: $K=0.28$). Table~\ref{tbl:humanVSmachine} shows accuracy against the gold labels, comparing humans (average accuracy over three annotators) to lexical and bleached models on the exact same subset of 200 users. Systems were tested under two different conditions regarding the number of tweets per user for the target language: machine and human saw the exact same twenty tweets, or the full set of tweets (200) per user, as done during training (Section~\ref{sec:models}). 

\begin{table}
\resizebox{\columnwidth}{!}{
	\begin{tabular}{l|  c| c c | c c }
    \toprule
    	  & Human & \multicolumn{2}{c}{Mach. \textsc{Lex}} & \multicolumn{2}{c}{Mach. \textsc{Abs}}\\
        tweets/user:    & 20          & 20 & 200 & 20 & 200  \\
              \midrule
				NL$\mapsto$NL & 70.5 & 69.0 & 81.0 & 49.5 & 72.0 \\
				NL$\mapsto$PT & 58.7 & 49.5 & 50.5 & 57.0 & 61.5 \\
				FR$\mapsto$NL & 60.3 & 50.0 & 50.0 & 50.5 & 62.0 \\
            \bottomrule
	\end{tabular}
    }
 \caption{Accuracy human versus machine.}
 \label{tbl:humanVSmachine}
\end{table}

First of all, our results indicate that in-language performance of humans is 70.5\%, which is quite in line with the  findings of \citet{flekova:etal:2016}, who report an accuracy of 75\% on English.  Within language, lexicalized models 
are superior to humans if exposed to enough information (200 tweets setup).  One explanation for this might lie in an observation by \citet{flekova:etal:2016}, according to which people tend to rely too much on stereotypical lexical indicators when assigning gender to the poster of a tweet, while machines model less evident patterns. Lexicalized models are also superior to the bleached ones, as already seen on the full datasets (Table~\ref{tbl:resultsMain}). 

We can also observe that the amount of information available to represent a user influences system's performance. Training on 200 tweets per user, but testing on 20 tweets only, decreases performance by 12 percentage points. This is likely due to the fact that inputs are sparser, especially since the bleached model is trained on 5-grams.\footnote{We experimented with training on 20 tweets rather than 200, and with different n-gram sizes (e.g., 1--4). Despite slightly better results, we decided to use the trained models as they were to employ the same settings  across all experiments (200 tweets per users, $n=5$), with no further tuning.}
The bleached model, when given 200 tweets per user, yields a performance that is slightly higher than human accuracy. 

In the cross-language setting, the picture is very different. Here, human performance is superior to the lexicalized models, independently of the amount of tweets per user at testing time. 
This seems to indicate that if  humans cannot rely on the lexicon, they might be exploiting some other signal when guessing the gender of a user who tweets in a language unknown to them. Interestingly, the bleached models, which rely on non-lexical features, not only outperform the lexicalized ones in the cross-language experiments, but also neatly match the human scores.

\section{Related Work}

Most existing work on gender prediction exploits shallow lexical information based on the linguistic production of the users. 
Few studies investigate deeper syntactic information~\cite{koppel2002,feng2012} or non-linguistic input, e.g., language-independent clues such as visual  \cite{alowibdi13-language} or network information  \cite{jurgens2013,plank-hovy:2015:WASSA,ljubevsic:ea:2017}. A related angle is cross-genre profiling. In both settings lexical models have limited portability due to their bias towards the language/genre they have been trained on~\cite{rangel16-overview,busger2016pan,medvedeva2017clef}. 

Lexical bias has been shown to affect in-language human gender prediction, too.  
\citet{flekova:etal:2016} found that people tend to rely too much on stereotypical lexical indicators, while  
\citet{nguyen2014gender} show that more than 10\% of the Twitter users do actually not employ words that the crowd associates with their biological sex.
Our features abstract away from such lexical cues while retaining predictive signal. 

\section{Conclusions}

Bleaching text into abstract features is surprisingly effective for predicting gender, though lexical information is still more useful within language (\textbf{RQ1}). However, models based on lexical clues fail when transferred to other languages, or require large amounts of unlabeled data from a similar domain as our experiments with the multilingual embedding model indicate. Instead, our bleached models clearly capture some signal beyond the lexicon, and perform well in a cross-lingual setting (\textbf{RQ2}). We are well aware that we are testing our cross-language bleached models in the context of closely related languages. While some features (such as PunctA, or Frequency) might carry over to genetically more distant languages, other features (such as Vowels and Shape) would probably be meaningless. Future work on this will require a sensible setting from a language typology perspective for choosing and testing adequate features.

In our novel study on human proficiency for cross-lingual gender prediction, we discovered that people  are also abstracting away from the lexicon. 
Indeed, we observe that they are able to detect gender by looking at tweets in a language they do not know  (\textbf{RQ3}) with an accuracy of 60\% on average. 

\section*{Acknowledgments}
We would like to thank the three anonymous reviewers and our colleagues for their useful feedback on earlier versions of this paper. Furthermore, we are grateful to Chlo\'{e} Braud for helping with the French human evaluation part. We would like to thank all of our human participants. 

\bibliography{main}
\bibliographystyle{acl_natbib}

\end{document}